\documentclass[sigconf, nonacm]{acmart}

\AtBeginDocument{%
}

\setcopyright{none}
\acmDOI{}
\acmISBN{}
\copyrightyear{}
\acmYear{}

\acmConference[SoCC '26]{ACM Symposium on Cloud Computing}{November 18--20, 2026}{Singapore}

\usepackage{tabularx}
\usepackage{array}
\usepackage{booktabs}
\usepackage{ragged2e}
\usepackage{xcolor}

\newcolumntype{Y}{>{\raggedright\arraybackslash}X}

\begin{document}
\thanks{Preprint. Submitted to ACM SoCC 2026.}

\title{The Rollout Infrastructure Tax in Coding-Agent Reinforcement Learning}

\author{Daniel Thi Graviet}
\affiliation{%
\institution{Daytona}
\country{United States}
}

\author{Lovre Pe\v{s}ut}
\affiliation{%
\institution{Daytona}
\country{United States}
}

\author{Ivan Dageli\'{c}}
\affiliation{%
\institution{Daytona}
\country{United States}
}

\author{Vedran Jukic}
\affiliation{%
\institution{Daytona}
\country{United States}
}

\author{Ivan Burazin}
\affiliation{%
\institution{Daytona}
\country{United States}
}

\renewcommand{\shortauthors}{Graviet et al.}


\begin{abstract}
Coding-agent reinforcement learning treats execution infrastructure as a background implementation detail, despite relying on large numbers of interactive software rollouts. This is a missed opportunity: measuring infrastructure overhead can reveal practical efficiency gains for RL post-training, where small per-rollout savings compound at scale. We present a comparative study of four execution substrates: single containers, hosted sandboxes, Kubernetes-orchestrated containers, and cloud virtual machines. We find up to 110$\times$ variation in cold-start latency and a 1.8$\times$ spread in projected worker-hours for one million 150-step trajectories. Our results suggest that future coding-agent RL systems should optimize execution substrates as part of the training system itself, not merely as deployment plumbing.
\end{abstract}

\ccsdesc[500]{Computer systems organization~Cloud computing}
\ccsdesc[300]{Software and its engineering~Software performance}
\ccsdesc[300]{Computing methodologies~Machine learning}

\maketitle

\section{Introduction}
Coding-agent reinforcement learning turns post-training into a large-scale systems workload. Instead of producing a single static answer, an agent repeatedly acts inside a software environment by inspecting files, editing code, running commands, and observing results. Each rollout therefore depends not only on model inference and reward computation, but also on the execution substrate that provisions and manages the environment.

This paper studies a specific systems question: how much does the execution substrate itself slow down coding-agent RL? The question is narrow, but consequential: seemingly small per-rollout and per-action delays compound quickly at training scale.

\begin{table}[htbp]
\centering
\caption{Small substrate overheads become large training costs.}
\label{tab:intro_scaling_example}
\begin{tabular}{lrr}
\toprule
\textbf{Extra overhead} & \textbf{Scale} & \textbf{Added worker-hours} \\
\midrule
1 s per rollout & 1M rollouts & 278 h \\
10 ms per action & 1M $\times$ 150 actions & 417 h \\
100 ms per action & 1M $\times$ 150 actions & 4{,}167 h \\
\bottomrule
\end{tabular}
\end{table}

Parallelism does not remove this cost; it converts it into a requirement for more workers, more scheduling capacity, and higher training expense to achieve the same wall-clock throughput.

We call this overhead the \emph{rollout infrastructure tax}: the latency and cost introduced by the systems that execute coding-agent RL trajectories. Execution substrates must provision environments, execute agent actions, manage state, and clean up after each attempt. They must also provide isolation and reproducibility so that one trajectory does not contaminate another or distort the reward signal. These requirements create tradeoffs among startup latency, per-action latency, scalability, and isolation.

Despite its importance, this tax is rarely measured directly. Recent agentic RL systems increasingly recognize rollout generation as a major training bottleneck, with some work reporting rollout generation as more than 90\% of total runtime~\cite{april}. Production traces further show that software environments introduce their own failures and long tails: containerized environment reset can dominate rollout time during failures, with delays caused by Docker image pulls, network contention, and host I/O contention~\cite{rollart}. These observations suggest that rollout execution is not merely a background implementation detail. However, existing work typically optimizes rollout scheduling, GPU utilization, or end-to-end training throughput, leaving open a narrower systems question: how much latency is introduced by the execution substrate itself?

We study this question through a measurement study of four coding-agent execution substrates: single containers, hosted sandboxes, Kubernetes-orchestrated containers, and cloud virtual machines. Our study makes the following contributions:

\begin{itemize}
\item We define the \emph{rollout infrastructure tax} and identify the systems operations that contribute to it, including environment creation, readiness, orchestration, command execution, observation collection, and cleanup (\S\ref{sec:rollout-tax}).

\item We present a controlled evaluation methodology for comparing execution substrates under identical coding-agent workloads (\S\ref{sec:evaluation-overview}).

\item We show that substrate choice directly affects rollout performance: cold-start latency varies by up to 110$\times$, but the gap narrows on heavier workloads (\S\ref{sec:results}).

\item We project these measurements to training scale and find that, for one million 150-step trajectories, substrate choice produces a 1.8$\times$ spread in rollout worker-hours, equivalent to 5{,}316 additional worker-hours (\S\ref{sec:training-scale}).

\item We translate the measurements into three concrete design requirements for rollout-native substrates: locality-aware warm pools, low-latency action APIs, and isolation mechanisms chosen according to rollout size (\S\ref{sec:design-implications}).
\end{itemize}

We do not claim that execution substrates solve RL post-training inefficiency, but show that they are a measurable and optimizable part of the training system. As coding-agent RL scales to larger workloads and longer trajectories, rollout infrastructure should be treated as a core systems concern rather than an implementation detail.

\section{The Rollout Infrastructure Tax}
\label{sec:rollout-tax}

The key idea in this paper is simple: a coding-agent rollout is not only a model computation; it is also a repeated interaction with an execution system. Before the agent can act, the system must create an environment and make it ready. Each time the agent acts, the system must execute a command, collect the result, update state, and return an observation. After the trajectory ends, the system must preserve, reset, or discard the environment so that later trials remain isolated and reproducible.

We call the latency introduced by this execution system the \emph{rollout infrastructure tax}. This tax excludes model inference and reward computation. It captures only the systems cost of running the environment in which the agent acts.

At a high level, a rollout has the following structure:

\begin{equation}
\underbrace{T_{\text{create}} + T_{\text{ready}}}_{\text{fixed per trajectory}}
+
\underbrace{\sum_{i=1}^{S} T_{\text{action},i}}_{\text{repeated per agent step}}
+
\underbrace{T_{\text{orchestration}}}_{\text{control-plane overhead}}
\label{eq:rollout-tax}
\end{equation}

where $S$ is the number of agent actions in the trajectory. This decomposition is useful because different parts of the tax matter in different regimes. For short trajectories, fixed startup and readiness costs can dominate total rollout time. For longer trajectories, those fixed costs are amortized, and repeated per-action overhead becomes increasingly important. A substrate can therefore look excellent for long rollouts but poor for short ones, or vice versa.

The components of the rollout infrastructure tax are:

\begin{itemize}
\item \textbf{Environment creation time.} Time required to provision a new execution environment.

\item \textbf{Readiness time.} Time between environment creation and the point at which the first agent action can execute. 

\item \textbf{Amortized per-action cost.} Total rollout latency divided by the number of agent actions.

\item \textbf{Orchestration overhead.} Latency introduced by the control plane that coordinates rollout execution. 
\end{itemize}

This decomposition gives the rest of the paper a measurement framework. Rather than asking only which substrate is fastest, we ask where each substrate pays the tax: before the first action, during each action, or in the control plane that coordinates the rollout. The answer determines when substrate choice matters most and how infrastructure overhead compounds at training scale.

\section{Evaluation Overview}
\label{sec:evaluation-overview}

We evaluate the rollout infrastructure tax by holding the coding-agent workload fixed while varying the execution substrate. Our goal is not to benchmark every possible deployment configuration, but to compare common substrate archetypes under a controlled rollout workload.

We study four common points in the rollout-substrate design space: lightweight single containers, commercial hosted coding-agent sandboxes, Kubernetes-orchestrated containers, and strongly isolated cloud virtual machines.

We evaluate three workload tiers. T0 is a minimal command workload that isolates startup overhead and the cost of launching commands. T1 uses a preloaded \texttt{python/attrs} repository to measure simple code inspection in an already-loaded repository, reflecting the code-navigation operations used by recent software-agent RL systems, including AST-based search tools~\cite{skyrlagent}. T2 uses \texttt{python/attrs} at a fixed buggy commit and includes repository cloning, source-code inspection, patch application, and test execution. This tier reflects the heavier setup and validation loop common in software-engineering agent tasks, where environment creation and step execution can become significant sources of latency~\cite{rollart}. Together, these tiers separate minimal substrate overhead from increasingly realistic coding-agent setup, execution, and validation costs.

Across these workloads, we measure cold-start latency, latency decomposition, multi-step trajectory completion time, and projected rollout worker-hours at training scale. Fresh-episode and multi-step trajectory experiments are each repeated 100 times per substrate and workload configuration. We report p50 and p95 latency to capture both typical and tail behavior.

All substrates use comparable software environments, resource limits, and regional placement when available. We avoid provider-specific optimizations such as specialized image prefetching, dependency caches, or scheduler tuning. The resulting measurements should be interpreted as controlled comparisons of execution substrates rather than theoretical limits of the underlying technologies.

\section{Results}
\label{sec:results}
We organize our evaluation around four research questions intended to characterize the rollout infrastructure tax and its implications for large-scale coding-agent reinforcement learning.

\subsection{RQ1: How Does Rollout Time Break Down?}

Our first goal is to understand how rollout latency is distributed across the execution lifecycle. Using the decomposition introduced in Section~\ref{sec:rollout-tax}, we measure environment creation latency, readiness latency, per-action execution latency, and orchestration overhead across all evaluated execution substrates.

This analysis reveals whether rollout performance is primarily constrained by startup costs, command responsiveness, scheduling delays, or workload setup. We focus on the T2 networked repository workload, which most closely resembles a realistic coding-agent trajectory: the environment clones a repository, performs source-code inspection, executes command-line operations, and resets state across repeated trials.

\begin{figure}[htbp]
\centering
\includegraphics[width=\columnwidth]{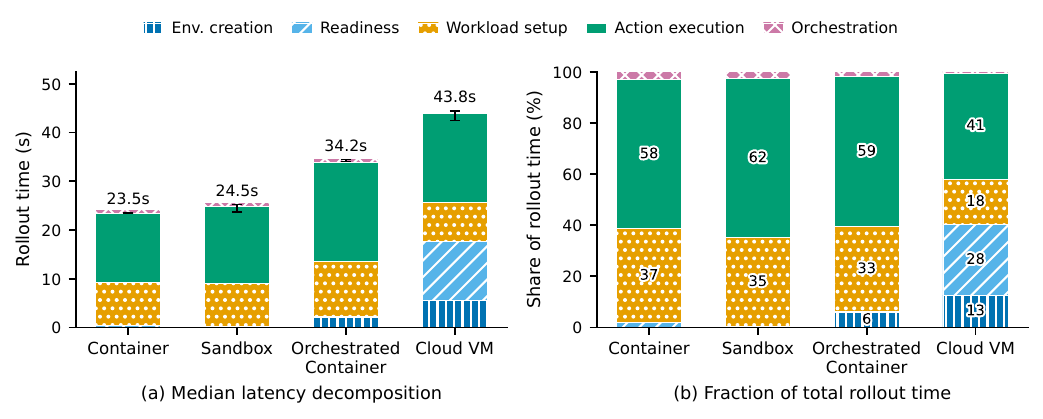}
\caption{Latency decomposition across execution substrates for Tier~2 workloads.}
\label{fig:rq1-latency-breakdown}
\end{figure}

As shown in Figure~\ref{fig:rq1-latency-breakdown}, the dominant source of latency varies across substrates. Lightweight container-based environments spend most of their rollout time on agent actions and setup, while substrates with stronger isolation or heavier orchestration incur larger startup and readiness costs. This result shows that rollout latency is not a single bottleneck: different substrates pay the infrastructure tax at different points in the execution lifecycle.

\subsection{RQ2: How Do Execution Substrates Compare Under Cold Starts?}

Many coding-agent RL systems provision a fresh execution environment for each trajectory, making cold-start latency a direct constraint on rollout throughput. This effect is especially pronounced for short-horizon tasks, where infrastructure overhead may dominate useful computation.

We measure cold-start latency across all substrates under identical workload and resource configurations. Each experiment provisions a new isolated environment and records the elapsed time until the first command can be executed. Results are reported for three workload tiers (T0--T2) using 100 trials per configuration.

\begin{figure}[htbp]
\centering
\includegraphics[width=\columnwidth]{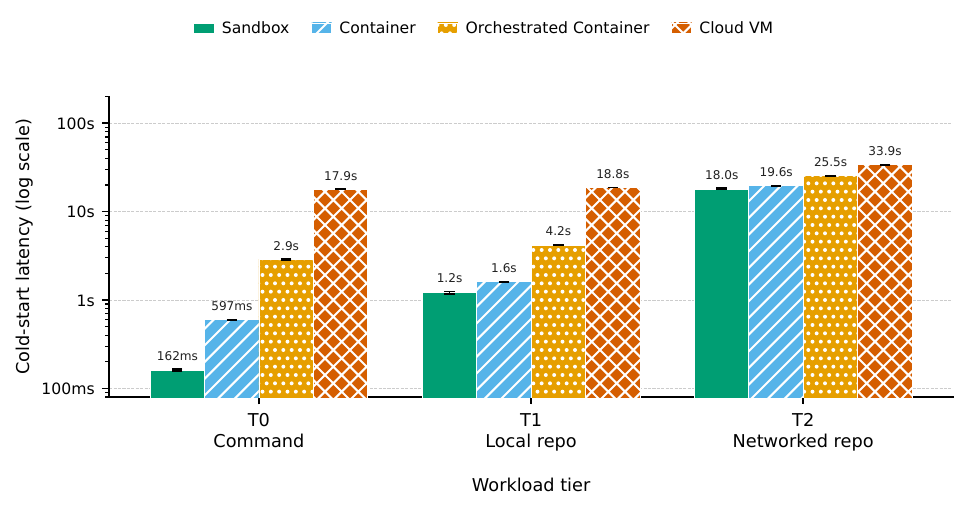}
\caption{P50 (median) cold-start latency by substrate and workload tier.}
\label{fig:rq2-coldstart}
\end{figure}

Figure~\ref{fig:rq2-coldstart} shows substantial variation in cold-start performance across execution substrates. The hosted sandbox achieves the lowest cold-start latency, followed by the single-container configuration. Kubernetes-orchestrated containers add scheduling overhead, while cloud VMs have the highest provisioning cost in our measured configuration. Across workload tiers, cold-start latency spans more than two orders of magnitude between the fastest and slowest substrates.

Cold-start overhead also affects lightweight workloads disproportionately. For T0 tasks, the gap between the hosted sandbox and the cloud VM exceeds two orders of magnitude. As workload complexity increases, however, cold-start latency is amortized by execution time: by T2, the cloud VM--hosted-sandbox gap shrinks to approximately 1.9$\times$.

\begin{figure}[t]
\centering
\includegraphics[width=\columnwidth]{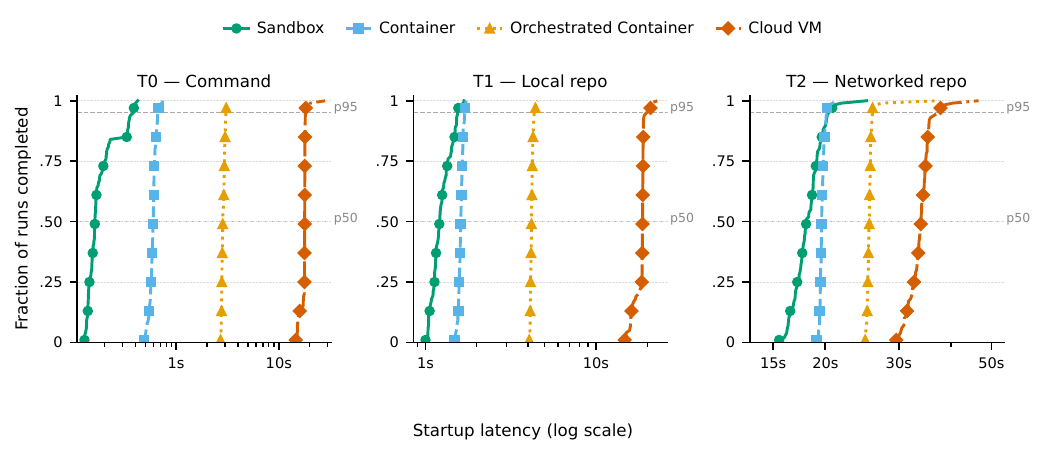}
\caption{Cold-start latency distribution across substrates.}
\label{fig:rq2-cdf}
\end{figure}

Figure~\ref{fig:rq2-cdf} shows the distributional behavior of cold starts. Orchestrated containers exhibit stable startup performance, with low variance and a narrow gap between median and tail latency. The hosted sandbox achieves the fastest median startup times but shows greater variability on lightweight workloads. Overall, these results show that substrate choice is most visible for short trajectories, where startup costs dominate, but remains important for longer rollouts because even smaller per-trajectory differences compound across parallel batches at training scale.

\subsection{RQ3: When Are Infrastructure Costs Amortized?}
\label{sec:amortization}

Cold-start latency alone does not determine rollout efficiency. Coding-agent RL trajectories often contain dozens or hundreds of agent actions, so fixed startup costs are divided across more actions as trajectories grow. We therefore measure how rollout latency and latency composition change with trajectory length.

We execute fixed-length trajectories containing 10, 50, and 150 actions. Each trajectory consists of representative coding-agent operations, including filesystem inspection, command execution, source-code modification, and test execution. We report both total trajectory completion time and amortized per-action latency, defined as total trajectory time divided by the number of actions. This second metric should be interpreted as an average cost per action including amortized startup and setup overhead, not as the latency of an isolated command.

\begin{figure}[htbp]
\centering
\includegraphics[width=\columnwidth]{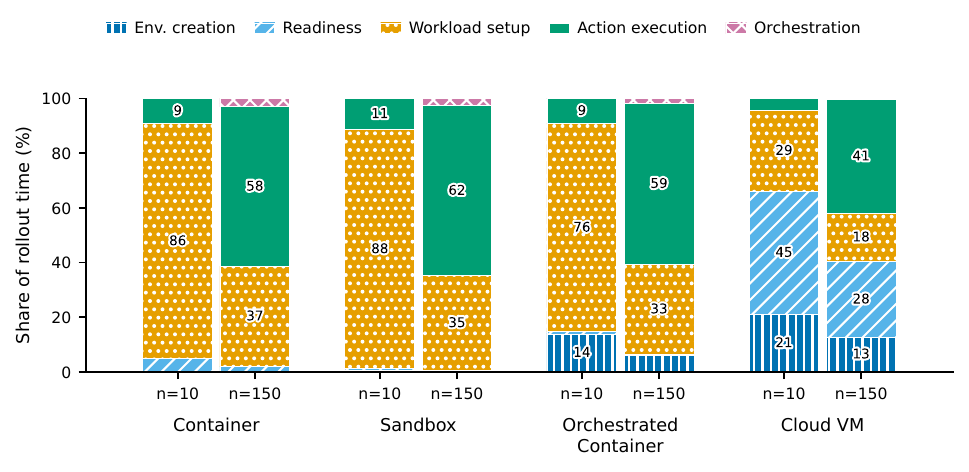}
\caption{Normalized latency composition as trajectories shift from 10 to 150 steps.}
\label{fig:rq3-component-shift}
\end{figure}

Figure~\ref{fig:rq3-component-shift} shows how the composition of rollout latency changes with trajectory length. Fixed infrastructure costs are most visible in short trajectories, while per-action execution becomes a larger share of total rollout time as trajectories grow.

\begin{figure}[t]
\centering
\includegraphics[width=\columnwidth]{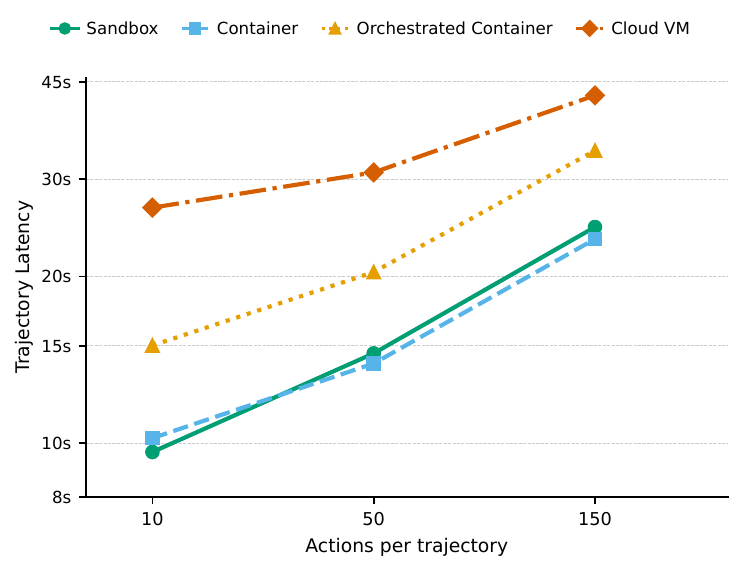}
\caption{Trajectory completion time by action count.}
\label{fig:rq3-trajectory-time}
\end{figure}

Figure~\ref{fig:rq3-trajectory-time} shows that increasing trajectory length amortizes fixed startup costs and narrows the relative gap between substrates. This does not mean that individual commands become faster. Rather, fixed provisioning and setup costs are divided across more actions. Short-horizon workloads amplify cold-start and readiness costs, while longer trajectories expose the ongoing cost of executing commands and returning observations at each agent step.

\subsection{RQ4: How Do Substrate Delays Compound at Training Scale?}
\label{sec:training-scale}

The previous results show that longer trajectories can amortize fixed startup costs. However, amortization does not make substrate overhead disappear. Even small per-trajectory latency gaps can become large operational costs when repeated across many rollout batches.

We estimate this effect by projecting measured 150-step trajectory latency into aggregate rollout worker-hours. This projection is complementary to recent work showing that rollout delays and long-tail trajectories can stall synchronous RL pipelines and leave accelerators underutilized~\cite{april,rollpacker}. In these settings, execution-substrate latency matters not only because rollout workers run longer, but also because delayed environment setup, action execution, and observation collection can slow the delivery of rollout data to downstream training components. We therefore focus on the substrate-side portion of this bottleneck: the time spent creating, preparing, and interacting with coding environments.

For each substrate $s$, let $L_s$ denote the measured end-to-end latency of a 150-step trajectory, excluding model inference and reward computation. For a training run containing $N$ trajectories, we estimate rollout worker-hours as
\begin{equation}
H_s(N) = \frac{N \cdot L_s}{3600}.
\label{eq:worker-hours}
\end{equation}

This models only rollout-worker demand, not total training cost. It excludes GPU inference, reward-model evaluation, optimizer steps, and queueing outside the rollout workers.

Table~\ref{tab:rq4-worker-hour-gap} shows how a small per-trajectory latency gap compounds at training scale.

\begin{table}[htbp]
\centering
\caption{A 19s per-trajectory gap compounds into thousands of worker-hours.}
\label{tab:rq4-worker-hour-gap}
\begin{tabular}{@{}lrr@{}}
\toprule
\textbf{Case} & \textbf{Latency / trajectory} & \textbf{Worker-hours} \\
\midrule
Slowest substrate & 42.5 s & 11{,}811 h \\
Fastest substrate & 23.4 s & 6{,}495 h \\
\midrule
Difference at 1M traj. & 19.1 s & 5{,}316 h \\
\bottomrule
\end{tabular}
\end{table}

Thus, a difference that is barely noticeable for a single rollout becomes a multi-thousand-worker-hour gap when repeated across one million 150-step trajectories.

\begin{figure}[htbp]
\centering
\includegraphics[width=\columnwidth]{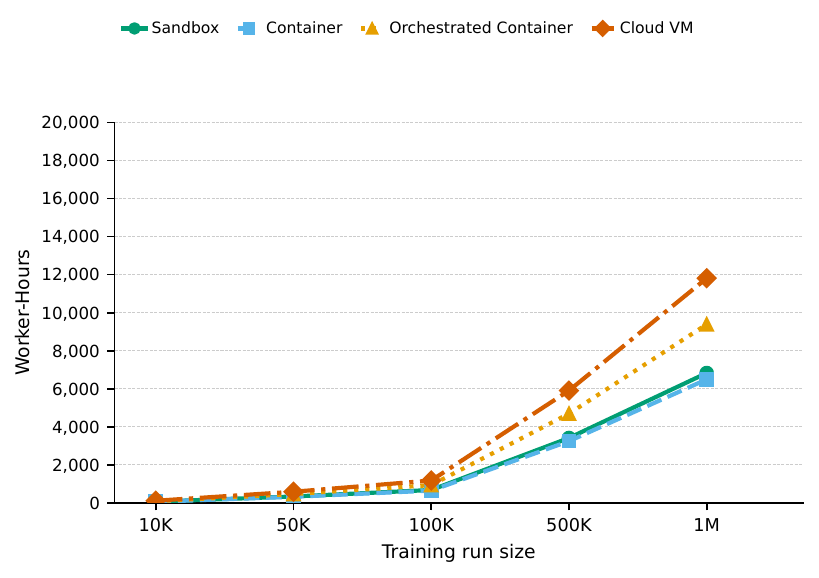}
\caption{Projected rollout worker-hours required at training scale across execution substrates.}
\label{fig:rq4-worker-hours}
\end{figure}

Figure~\ref{fig:rq4-worker-hours} shows this effect across training runs from 10{,}000 to 1{,}000{,}000 trajectories. As rollout counts grow, the same per-trajectory latency gap turns into a larger worker-hour gap.

\section{Design Implications}
\label{sec:design-implications}

Our results suggest that substrates designed for long-lived services, batch jobs, or general-purpose cloud workloads can expose overheads that become costly when repeated across many short, interactive trajectories. This motivates rollout-native designs that reduce trajectory latency while preserving isolation, reproducibility, and scalable execution. We highlight three implications.

\emph{Move readiness work off the critical path with locality-aware warm pools.} Short trajectories are especially sensitive to startup and readiness latency because fixed costs are amortized over only a small number of actions. Rollout systems should therefore avoid repeatedly constructing environments from scratch. A rollout-native substrate can maintain staged warm pools: base environments with the runtime already initialized, image-warmed environments with required tools loaded, repository-warmed environments with source trees already present, and dependency-warmed environments with language packages or build artifacts cached. These pools are most effective when the scheduler accounts for the state already present in each environment. Rather than asking only whether a worker is idle, the scheduler should ask which idle worker is closest to the desired rollout state. Routing a trajectory to a worker with the relevant image, repository, dependency cache, or filesystem state can reduce readiness costs before useful agent work begins.

\emph{Minimize control-plane and command-dispatch overhead.} Rollout execution is interactive: every trajectory may require tens or hundreds of commands, observations, file operations, and test executions. Substrates that route each action through remote shells, orchestration APIs, scheduling layers, or heavyweight result-collection paths can add latency repeatedly across the trajectory. As trajectories grow longer and startup costs are amortized, low-latency action APIs, direct process execution, fast filesystem access, and lightweight observation collection become increasingly important.

\emph{Match isolation and scaling mechanisms to rollout economics.} Strong isolation and elastic scaling are necessary for large-scale rollout generation, but they can impose startup, reset, scheduling, and placement costs. Virtual machines provide strong isolation, but their fresh-start paths can be expensive. Containers reduce startup overhead, but may provide weaker isolation. MicroVMs, snapshots, sandbox runtimes, and preallocated rollout workers offer intermediate designs. The appropriate substrate depends on rollout length, reset frequency, parallelism, and the cost of waiting for new workers to become ready.

Taken together, these implications point toward execution substrates that combine warm reuse, cache locality, low-latency interaction, scalable scheduling, and lightweight isolation. Treating coding-agent rollouts as a distinct systems workload creates an opportunity to reduce training cost without changing the model, reward function, or optimization algorithm.

\section{Limitations}

This study is a controlled measurement of rollout infrastructure rather than a complete simulation of production coding-agent RL systems. Our workloads use fixed command sequences to make comparisons reproducible across substrates; real agents may issue more varied commands, interact with larger repositories, or exhibit longer-horizon behavior. However, our study captures the core operations common to coding-agent rollouts: environment creation, command execution, filesystem interaction, test execution, and reset.

Our measurements also reflect specific hardware, regions, cache states, and provider configurations. Cloud execution latency can vary with image locality, resource placement, noisy neighbors, storage performance, and control-plane load. We therefore interpret the absolute latencies as configuration-specific, while treating the cross-substrate differences and scaling trends as the primary result.

Finally, our training-scale projections depend on assumptions about trajectory length, rollout count, parallelism, and environment reuse. These projections are not intended as universal cost estimates. Instead, they show how measured per-trajectory overheads compound at scale and provide a comparative framework for reasoning about rollout infrastructure choices.

\section{Related Work}

\textbf{Coding agents and interactive software environments.}
Recent work on coding agents has moved language models from single-turn code generation into interactive software environments, where agents inspect repositories, edit files, execute commands, and run tests. Benchmarks and systems such as TerminalBench, SWE-bench, and SWE-agent have been especially valuable in making this interaction loop concrete~\cite{tbench,swebench,sweagent}. Our work studies the same execution pattern of stateful, multi-step trajectories over software repositories. However, it asks a different question: rather than measuring agent success rate or task realism, we measure the infrastructure cost of executing these trajectories at training scale.

\textbf{Reinforcement learning and rollout systems.}
Reinforcement learning from feedback and reasoning-oriented post-training methods rely on large numbers of sampled trajectories or rollouts~\cite{rlhf,deepseekr1}. Distributed systems such as Ray and RLlib show how rollout generation and task execution can be scaled across clusters~\cite{ray,rllibflow}. This line of work motivates our focus on rollout throughput. However, coding-agent RL adds an execution loop that is not present in many standard RL settings: each trajectory requires a mutable software environment that must be provisioned, prepared, interacted with, and reset. Our study isolates this environment-execution cost rather than the model, reward, or optimization components of RL training.

\textbf{Execution substrates, cold starts, and ML infrastructure.}
Systems research has developed techniques for reducing startup latency, improving isolation, and scheduling distributed workloads. Firecracker, SEUSS, and Catalyzer show how lightweight virtualization, snapshots, and runtime initialization affect startup latency and isolation~\cite{firecracker,cadden2020seuss,du2020catalyzer}. Cluster and ML systems such as Borg, Ray, and Tiresias study scheduling, resource allocation, and training throughput at scale~\cite{borg,ray,gu2019tiresias}. We build on these systems insights, but apply them to repeated, interactive, stateful coding-agent rollouts. In this setting, cold starts, command execution, filesystem mutation, observation collection, and environment reset all become part of the training loop.

\section{Conclusion}

Coding-agent RL makes execution infrastructure part of the training loop. Each rollout must create, prepare, interact with, and observe a software environment, and these costs accumulate across large numbers of trajectories.

This paper measured that cost as the \emph{rollout infrastructure tax}. Across four execution substrates, we showed that substrate choice can change cold-start latency by orders of magnitude and produce thousands of additional rollout worker-hours at training scale. The main lesson is not that one substrate is universally best, but that rollout infrastructure is measurable, consequential, and optimizable.

Future coding-agent RL systems should treat execution substrates as part of the training system itself. Rollout-native stacks that combine warm reuse, cache locality, low-latency interaction, efficient reset, scalable orchestration, and appropriate isolation may reduce training cost without changing the model, reward function, or optimization algorithm.

\bibliographystyle{ACM-Reference-Format}
\bibliography{software}

@misc{tbench,
      title={Terminal-Bench: Benchmarking Agents on Hard, Realistic Tasks in Command Line Interfaces}, 
      author={Mike A. Merrill and Alexander G. Shaw and Nicholas Carlini and Boxuan Li and Harsh Raj and Ivan Bercovich and Lin Shi and Jeong Yeon Shin and Thomas Walshe and E. Kelly Buchanan and Junhong Shen and Guanghao Ye and Haowei Lin and Jason Poulos and Maoyu Wang and Marianna Nezhurina and Jenia Jitsev and Di Lu and Orfeas Menis Mastromichalakis and Zhiwei Xu and Zizhao Chen and Yue Liu and Robert Zhang and Leon Liangyu Chen and Anurag Kashyap and Jan-Lucas Uslu and Jeffrey Li and Jianbo Wu and Minghao Yan and Song Bian and Vedang Sharma and Ke Sun and Steven Dillmann and Akshay Anand and Andrew Lanpouthakoun and Bardia Koopah and Changran Hu and Etash Guha and Gabriel H. S. Dreiman and Jiacheng Zhu and Karl Krauth and Li Zhong and Niklas Muennighoff and Robert Amanfu and Shangyin Tan and Shreyas Pimpalgaonkar and Tushar Aggarwal and Xiangning Lin and Xin Lan and Xuandong Zhao and Yiqing Liang and Yuanli Wang and Zilong Wang and Changzhi Zhou and David Heineman and Hange Liu and Harsh Trivedi and John Yang and Junhong Lin and Manish Shetty and Michael Yang and Nabil Omi and Negin Raoof and Shanda Li and Terry Yue Zhuo and Wuwei Lin and Yiwei Dai and Yuxin Wang and Wenhao Chai and Shang Zhou and Dariush Wahdany and Ziyu She and Jiaming Hu and Zhikang Dong and Yuxuan Zhu and Sasha Cui and Ahson Saiyed and Arinbjörn Kolbeinsson and Jesse Hu and Christopher Michael Rytting and Ryan Marten and Yixin Wang and Alex Dimakis and Andy Konwinski and Ludwig Schmidt},
      year={2026},
      eprint={2601.11868},
      archivePrefix={arXiv},
      primaryClass={cs.SE},
      url={https://arxiv.org/abs/2601.11868}, 
}

@misc{swebench,
      title={SWE-bench: Can Language Models Resolve Real-World GitHub Issues?}, 
      author={Carlos E. Jimenez and John Yang and Alexander Wettig and Shunyu Yao and Kexin Pei and Ofir Press and Karthik Narasimhan},
      year={2024},
      eprint={2310.06770},
      archivePrefix={arXiv},
      primaryClass={cs.CL},
      url={https://arxiv.org/abs/2310.06770}, 
}

@misc{sweagent,
      title={SWE-agent: Agent-Computer Interfaces Enable Automated Software Engineering}, 
      author={John Yang and Carlos E. Jimenez and Alexander Wettig and Kilian Lieret and Shunyu Yao and Karthik Narasimhan and Ofir Press},
      year={2024},
      eprint={2405.15793},
      archivePrefix={arXiv},
      primaryClass={cs.SE},
      url={https://arxiv.org/abs/2405.15793}, 
}

@misc{rlhf,
      title={Training language models to follow instructions with human feedback}, 
      author={Long Ouyang and Jeff Wu and Xu Jiang and Diogo Almeida and Carroll L. Wainwright and Pamela Mishkin and Chong Zhang and Sandhini Agarwal and Katarina Slama and Alex Ray and John Schulman and Jacob Hilton and Fraser Kelton and Luke Miller and Maddie Simens and Amanda Askell and Peter Welinder and Paul Christiano and Jan Leike and Ryan Lowe},
      year={2022},
      eprint={2203.02155},
      archivePrefix={arXiv},
      primaryClass={cs.CL},
      url={https://arxiv.org/abs/2203.02155}, 
}

@article{deepseekr1,
   title={DeepSeek-R1 incentivizes reasoning in LLMs through reinforcement learning},
   volume={645},
   ISSN={1476-4687},
   url={http://dx.doi.org/10.1038/s41586-025-09422-z},
   DOI={10.1038/s41586-025-09422-z},
   number={8081},
   journal={Nature},
   publisher={Springer Science and Business Media LLC},
   author={Guo, Daya and Yang, Dejian and Zhang, Haowei and Song, Junxiao and Wang, Peiyi and Zhu, Qihao and Xu, Runxin and Zhang, Ruoyu and Ma, Shirong and Bi, Xiao and Zhang, Xiaokang and Yu, Xingkai and Wu, Yu and Wu, Z. F. and Gou, Zhibin and Shao, Zhihong and Li, Zhuoshu and Gao, Ziyi and Liu, Aixin and Xue, Bing and Wang, Bingxuan and Wu, Bochao and Feng, Bei and Lu, Chengda and Zhao, Chenggang and Deng, Chengqi and Ruan, Chong and Dai, Damai and Chen, Deli and Ji, Dongjie and Li, Erhang and Lin, Fangyun and Dai, Fucong and Luo, Fuli and Hao, Guangbo and Chen, Guanting and Li, Guowei and Zhang, H. and Xu, Hanwei and Ding, Honghui and Gao, Huazuo and Qu, Hui and Li, Hui and Guo, Jianzhong and Li, Jiashi and Chen, Jingchang and Yuan, Jingyang and Tu, Jinhao and Qiu, Junjie and Li, Junlong and Cai, J. L. and Ni, Jiaqi and Liang, Jian and Chen, Jin and Dong, Kai and Hu, Kai and You, Kaichao and Gao, Kaige and Guan, Kang and Huang, Kexin and Yu, Kuai and Wang, Lean and Zhang, Lecong and Zhao, Liang and Wang, Litong and Zhang, Liyue and Xu, Lei and Xia, Leyi and Zhang, Mingchuan and Zhang, Minghua and Tang, Minghui and Zhou, Mingxu and Li, Meng and Wang, Miaojun and Li, Mingming and Tian, Ning and Huang, Panpan and Zhang, Peng and Wang, Qiancheng and Chen, Qinyu and Du, Qiushi and Ge, Ruiqi and Zhang, Ruisong and Pan, Ruizhe and Wang, Runji and Chen, R. J. and Jin, R. L. and Chen, Ruyi and Lu, Shanghao and Zhou, Shangyan and Chen, Shanhuang and Ye, Shengfeng and Wang, Shiyu and Yu, Shuiping and Zhou, Shunfeng and Pan, Shuting and Li, S. S. and Zhou, Shuang and Wu, Shaoqing and Yun, Tao and Pei, Tian and Sun, Tianyu and Wang, T. and Zeng, Wangding and Liu, Wen and Liang, Wenfeng and Gao, Wenjun and Yu, Wenqin and Zhang, Wentao and Xiao, W. L. and An, Wei and Liu, Xiaodong and Wang, Xiaohan and Chen, Xiaokang and Nie, Xiaotao and Cheng, Xin and Liu, Xin and Xie, Xin and Liu, Xingchao and Yang, Xinyu and Li, Xinyuan and Su, Xuecheng and Lin, Xuheng and Li, X. Q. and Jin, Xiangyue and Shen, Xiaojin and Chen, Xiaosha and Sun, Xiaowen and Wang, Xiaoxiang and Song, Xinnan and Zhou, Xinyi and Wang, Xianzu and Shan, Xinxia and Li, Y. K. and Wang, Y. Q. and Wei, Y. X. and Zhang, Yang and Xu, Yanhong and Li, Yao and Zhao, Yao and Sun, Yaofeng and Wang, Yaohui and Yu, Yi and Zhang, Yichao and Shi, Yifan and Xiong, Yiliang and He, Ying and Piao, Yishi and Wang, Yisong and Tan, Yixuan and Ma, Yiyang and Liu, Yiyuan and Guo, Yongqiang and Ou, Yuan and Wang, Yuduan and Gong, Yue and Zou, Yuheng and He, Yujia and Xiong, Yunfan and Luo, Yuxiang and You, Yuxiang and Liu, Yuxuan and Zhou, Yuyang and Zhu, Y. X. and Huang, Yanping and Li, Yaohui and Zheng, Yi and Zhu, Yuchen and Ma, Yunxian and Tang, Ying and Zha, Yukun and Yan, Yuting and Ren, Z. Z. and Ren, Zehui and Sha, Zhangli and Fu, Zhe and Xu, Zhean and Xie, Zhenda and Zhang, Zhengyan and Hao, Zhewen and Ma, Zhicheng and Yan, Zhigang and Wu, Zhiyu and Gu, Zihui and Zhu, Zijia and Liu, Zijun and Li, Zilin and Xie, Ziwei and Song, Ziyang and Pan, Zizheng and Huang, Zhen and Xu, Zhipeng and Zhang, Zhongyu and Zhang, Zhen},
   year={2025},
   month=Sept, pages={633–638} }

@misc{ray,
      title={Ray: A Distributed Framework for Emerging AI Applications}, 
      author={Philipp Moritz and Robert Nishihara and Stephanie Wang and Alexey Tumanov and Richard Liaw and Eric Liang and Melih Elibol and Zongheng Yang and William Paul and Michael I. Jordan and Ion Stoica},
      year={2018},
      eprint={1712.05889},
      archivePrefix={arXiv},
      primaryClass={cs.DC},
      url={https://arxiv.org/abs/1712.05889}, 
}

@misc{rllibflow,
      title={RLlib Flow: Distributed Reinforcement Learning is a Dataflow Problem}, 
      author={Eric Liang and Zhanghao Wu and Michael Luo and Sven Mika and Joseph E. Gonzalez and Ion Stoica},
      year={2021},
      eprint={2011.12719},
      archivePrefix={arXiv},
      primaryClass={cs.LG},
      url={https://arxiv.org/abs/2011.12719}, 
}

@inproceedings {firecracker,
author = {Alexandru Agache and Marc Brooker and Alexandra Iordache and Anthony Liguori and Rolf Neugebauer and Phil Piwonka and Diana-Maria Popa},
title = {Firecracker: Lightweight Virtualization for Serverless Applications },
booktitle = {17th USENIX Symposium on Networked Systems Design and Implementation (NSDI 20)},
year = {2020},
isbn = {978-1-939133-13-7},
address = {Santa Clara, CA},
pages = {419--434},
url = {https://www.usenix.org/conference/nsdi20/presentation/agache},
publisher = {USENIX Association},
month = feb
}

@inproceedings{cadden2020seuss,
title = {{SEUSS}: Skip Redundant Paths to Make Serverless Fast},
author = {Cadden, James and Unger, Thomas and Awad, Yara and Dong, Han and Krieger, Orran and Appavoo, Jonathan},
booktitle = {Proceedings of the Fifteenth European Conference on Computer Systems},
year = {2020},
articleno = {32},
pages = {1--15},
doi = {10.1145/3342195.3392698},
url = {https://doi.org/10.1145/3342195.3392698}
}

@inproceedings{du2020catalyzer,
title = {Catalyzer: Sub-millisecond Startup for Serverless Computing with Initialization-less Booting},
author = {Du, Dong and Yu, Tianyi and Xia, Yubin and Zang, Binyu and Yan, Guanglu and Qin, Chenggang and Wu, Qixuan and Chen, Haibo},
booktitle = {Proceedings of the Twenty-Fifth International Conference on Architectural Support for Programming Languages and Operating Systems},
year = {2020},
pages = {467--481},
doi = {10.1145/3373376.3378512},
url = {https://doi.org/10.1145/3373376.3378512}
}

@article{borg,title	= {Borg, Omega, and Kubernetes},author	= {Brendan Burns and Brian Grant and David Oppenheimer and Eric Brewer and John Wilkes},year	= {2016},URL	= {http://queue.acm.org/detail.cfm?id=2898444},journal	= {ACM Queue},pages	= {70--93},volume	= {14}}

@inproceedings{gu2019tiresias,
title = {Tiresias: A {GPU} Cluster Manager for Distributed Deep Learning},
author = {Gu, Juncheng and Chowdhury, Mosharaf and Shin, Kang G. and Zhu, Yibo and Jeon, Myeongjae and Qian, Junjie and Liu, Hongqiang Harry and Guo, Chuanxiong},
booktitle = {16th USENIX Symposium on Networked Systems Design and Implementation},
year = {2019},
pages = {485--500},
url = {https://www.usenix.org/conference/nsdi19/presentation/gu}
}

@misc{april,
      title={APRIL: Active Partial Rollouts in Reinforcement Learning to Tame Long-tail Generation}, 
      author={Yuzhen Zhou and Jiajun Li and Yusheng Su and Gowtham Ramesh and Zilin Zhu and Xiang Long and Chenyang Zhao and Jin Pan and Xiaodong Yu and Ze Wang and Kangrui Du and Jialian Wu and Ximeng Sun and Jiang Liu and Qiaolin Yu and Hao Chen and Zicheng Liu and Emad Barsoum},
      year={2025},
      eprint={2509.18521},
      archivePrefix={arXiv},
      primaryClass={cs.LG},
      url={https://arxiv.org/abs/2509.18521}, 
}

@misc{rollart,
      title={RollArt: Disaggregated Multi-Task Agentic RL Training at Scale}, 
      author={Wei Gao and Yuheng Zhao and Tianyuan Wu and Shaopan Xiong and Weixun Wang and Dakai An and Lunxi Cao and Dilxat Muhtar and Zichen Liu and Haizhou Zhao and Ju Huang and Siran Yang and Yongbin Li and Wenbo Su and Jiamang Wang and Lin Qu and Bo Zheng and Wei Wang},
      year={2026},
      eprint={2512.22560},
      archivePrefix={arXiv},
      primaryClass={cs.DC},
      url={https://arxiv.org/abs/2512.22560}, 
}

@misc{skyrlagent,
      title={SkyRL-Agent: Efficient RL Training for Multi-turn LLM Agent}, 
      author={Shiyi Cao and Dacheng Li and Fangzhou Zhao and Shuo Yuan and Sumanth R. Hegde and Connor Chen and Charlie Ruan and Tyler Griggs and Shu Liu and Eric Tang and Richard Liaw and Philipp Moritz and Matei Zaharia and Joseph E. Gonzalez and Ion Stoica},
      year={2025},
      eprint={2511.16108},
      archivePrefix={arXiv},
      primaryClass={cs.AI},
      url={https://arxiv.org/abs/2511.16108}, 
}

@misc{rollpacker,
      title={RollPacker: Mitigating Long-Tail Rollouts for Fast, Synchronous RL Post-Training}, 
      author={Wei Gao and Yuheng Zhao and Dakai An and Tianyuan Wu and Lunxi Cao and Shaopan Xiong and Ju Huang and Weixun Wang and Siran Yang and Wenbo Su and Jiamang Wang and Lin Qu and Bo Zheng and Wei Wang},
      year={2025},
      eprint={2509.21009},
      archivePrefix={arXiv},
      primaryClass={cs.DC},
      url={https://arxiv.org/abs/2509.21009}, 
}

\end{document}